\PassOptionsToPackage{table}{xcolor}
\documentclass[sigconf]{acmart}
\AtBeginDocument{%
  }

\setcopyright{acmlicensed}
\copyrightyear{2018}
\acmYear{2018}
\acmDOI{XXXXXXX.XXXXXXX}
\acmConference[Conference acronym 'XX]{Make sure to enter the correct
  conference title from your rights confirmation email}{June 03--05,
  2018}{Woodstock, NY}
\acmISBN{978-1-4503-XXXX-X/2018/06}

\usepackage{amsmath}
\usepackage{nicefrac}
\usepackage{algorithm}
\usepackage{algorithmic}

\usepackage{multirow}
\usepackage{makecell}
\usepackage{subcaption}

\begin{document}

\title{APO2CS: Advantage-Aware Preference Optimization for Two-Level Constructive Search in Combinatorial Optimization}


\author{Kaixi Yang}
\affiliation{%
  \institution{National University of Defense Technology}
  \city{Changsha}
  \country{China}}
\email{kxyang1998@nudt.edu.cn}

\author{Linbo Qiao}
\correspondingauthor
\affiliation{%
  \institution{National University of Defense Technology}
  \city{Changsha}
  \country{China}
}
\email{qiao.linbo@nudt.edu.cn}

\author{Wei Shang}
\affiliation{%
  \institution{National University of Defense Technology}
  \city{Changsha}
  \country{China}}
\email{shangwei24@nudt.edu.cn}

\author{Yanqi Shi}
\affiliation{%
  \institution{National University of Defense Technology}
  \city{Changsha}
  \country{China}}
\email{yqshi@nudt.edu.cn}

\author{Dongsheng Li}
\affiliation{%
  \institution{National University of Defense Technology}
  \city{Changsha}
  \country{China}}
\email{dsli@nudt.edu.cn}

\renewcommand{\shortauthors}{Trovato et al.}

\begin{abstract}
Preference optimization has emerged as a promising alternative to reward-based training for neural combinatorial optimization. However, existing methods typically rely on binary pairwise labels, which discard the magnitude of solution-quality differences and provide supervision only at the complete-solution level. We propose Advantage-aware Preference Optimization for Two-level Constructive Search (APO2CS), a preference-based training objective for constructive neural solvers that integrates both macro-level and micro-level supervision. At the macro level, APO2CS enriches pairwise preferences with an advantage term derived from objective-value differences. At the micro level, it compares partial trajectories, thereby providing denser supervision throughout the sequential construction process. The resulting objective can be used either to train a policy from scratch or to fine-tune a pretrained one. Experimental results show that APO2CS improves the expressiveness and search capability of preference learning for combinatorial optimization, while demonstrating strong effectiveness and adaptability across multiple combinatorial optimization tasks.
\end{abstract}

\ccsdesc[500]{Computing methodologies~Artificial intelligence}
\ccsdesc[300]{Computing methodologies~Machine learning}
\ccsdesc[100]{Computing methodologies~Reinforcement learning}
\keywords{Combinatorial Optimization, Learning to Optimize, Neural Combinatorial Optimization, Preference Optimization, Reinforcement Learning}

\received{20 February 2007}
\received[revised]{12 March 2009}
\received[accepted]{5 June 2009}

\maketitle

\section{Introduction}
Combinatorial optimization problems (COPs) \cite{papadimitriou1998combinatorial} arise extensively in practical applications. In general, their goal is to determine a solution that minimizes or maximizes an objective function over a finite but typically exponentially large solution space, subject to a set of constraints. Classic examples of COPs include the Traveling Salesman Problem (TSP)\cite{tsp}, the Vehicle Routing Problem (VRP)\cite{vrp}, the Job Shop Scheduling Problem (JSSP)\cite{jssp}, and the Knapsack Problem (KP)\cite{kp}. Beyond their fundamental theoretical significance, these problems are also central to numerous real-world applications, such as logistics and distribution \cite{waters2003global}, communication network design \cite{hrasnica2005broadband}, chip routing \cite{rantala2006network}, resource allocation \cite{katoh1998resource}, and production scheduling \cite{graves1981review}.

In contrast to continuous optimization problems, COPs are typically characterized by highly non-convex and non-smooth solution spaces, posing significant challenges to gradient-based optimization methods. Furthermore, many classical COPs are NP-hard, making it generally impractical to obtain global optimal solutions within an acceptable computational time. Therefore, from the perspective of real-world applications, the efficient acquisition of high-quality approximate solutions is of paramount importance.

With the advancement of deep learning \cite{lecun2015deep}, the paradigm for solving COPs has progressively shifted from traditional exact and heuristic algorithms toward learning-based approximate approaches. Conventional supervised learning \cite{cunningham2008supervised} typically relies on precise labels or objective values as training signals. However, in COPs, such supervision is often expensive to obtain and susceptible to numerical noise, making it difficult for models to effectively capture the structural differences in solution quality. In contrast, preference optimization (PO) \cite{ucpo}\cite{bopo}\cite{poco} replaces absolute objective supervision with relative comparisons between solutions, making it a more suitable paradigm when high-quality explicit supervision is difficult to acquire. Despite this advantage, most existing PO methods rely on discretized binary preference labels. While this design improves training stability, it inevitably compresses the underlying optimization information and limits the model’s ability to capture the magnitude of advantage between solutions. Moreover, these methods are often highly sensitive to the quality of the candidate solution set, with their learning dynamics largely constrained by the sample distribution within the candidate pool, effectively reducing the task to posterior re-ranking over a fixed set of candidates. This limitation, in turn, restricts the broader generalizability of PO.

To address these limitations, we propose a novel PO algorithm, APO2CS, that integrates advantage-aware and two-level constructive search. Our contributions are:
\begin{itemize}
    \item We introduce an advantage-aware preference objective that incorporates pairwise objective differences into preference-based training for constructive neural solvers.
    \item We propose a two-level supervision scheme that combines macro-level and micro-level preference learning, enabling global ranking and local trajectory refinement to be optimized jointly. 
    \item We show that the resulting objective can be applied in both training-from-scratch and fine-tuning settings, and we empirically study its behavior on benchmarks.
\end{itemize}

\section{Related Work}
\paragraph{Learning-based Combinatorial Optimization.} Recent years have witnessed rapid progress in learning-based combinatorial optimization, where neural models are trained to directly construct or improve solutions. Early work, such as Pointer Networks \cite{pointer}, demonstrated that sequence-to-sequence architectures with attention \cite{vaswani2017attention} can effectively model combinatorial output spaces and solve problems such as the TSP. Building on this line of research, \cite{bello2017neural} introduced reinforcement learning (RL) \cite{shakya2023reinforcement} to directly optimize solution quality, showing that neural policies can learn heuristic solvers without handcrafted rules. Subsequent studies further improved model architectures and training strategies. For instance, \cite{kool2019attentionlearnsolverouting} proposed an attention-based autoregressive solver that has become a representative framework for neural routing optimization. Hybrid approaches such as NeuroLKH \cite{xin2021neurolkh} further demonstrated that learned modules can be effectively integrated with strong traditional heuristics, leading to substantial improvements in both solution quality and computational efficiency. However, existing learning-based methods still rely heavily on explicit labels, scalar rewards, or handcrafted solver integration, which limits their robustness and applicability when high-quality supervision is costly or difficult to design. More recently, inspired by the success of preference-based alignment methods such as Direct Preference Optimization (DPO) \cite{dpo}, an emerging line of research has begun to replace explicit scalar supervision with relative preference signals between candidate solutions, opening a new avenue for PO in combinatorial optimization.

\paragraph{Preference Optimization.} Preference Optimization offers an alternative supervision paradigm by replacing absolute objective values with relative comparisons between candidate solutions. Classical studies on ranking from pairwise preferences, such as \cite{ailon2010preferencebased} and \cite{ailon2012active}, established the fundamental formulation of learning from pairwise preference labels and highlighted the effectiveness of relative supervision for ranking tasks. These studies laid the theoretical foundation for applying PO to combinatorial optimization. Building on this foundation, recent methods such as POCO \cite{poco} and BOPO \cite{bopo} construct preference pairs from objective values and employ preference-based objectives to improve exploration and optimization efficiency. More recently, UCPO \cite{ucpo} extended this line of research to constrained combinatorial optimization by incorporating constraint satisfaction directly into the preference objective, thereby providing a plug-and-play framework for handling complex constraints without requiring architectural modifications. Nevertheless, most existing PO methods rely on binary preference labels, which capture only relative ordering while failing to reflect the magnitude of superiority between solutions.

\paragraph{Constructive Search.} Another important line of research focuses on constructive search, where solutions are generated sequentially through a series of decisions. Early neural solvers, as well as later attention-based routing models, adopt an autoregressive construction process that enables the model to incrementally build solutions rather than merely score complete candidates after generation. In a similar spirit, POMO \cite{pomo} improves constructive RL by exploiting multiple optima and encouraging diverse rollout trajectories, thereby enhancing both training stability and solution quality. Beyond purely constructive generation, NeuRewriter \cite{chen2019learning} demonstrates that search can be combined with learned local rewriting policies, enabling iterative refinement of intermediate solutions during optimization. Despite their effectiveness, existing constructive search methods typically optimize generation through rewards or heuristics alone, without explicitly incorporating fine-grained preference signals into both local trajectory construction and global solution evaluation.

Our work bridges these gaps by introducing finer-grained preference comparison and a two-level solution construction strategy within the PO framework. Specifically, APO2CS explicitly models both preference direction and advantage magnitude, while incorporating a two-level constructive search mechanism that injects preference signals into both local trajectory optimization and global solution comparison.

\section{Motivation}
\label{motivation}
Most existing preference optimization methods for combinatorial optimization rely on binary pairwise preference labels. Given a COP instance $x \in \mathcal{X}$, the objective is to solve
\begin{equation}
    \min_{s \in \mathcal{S}(x)} f(x, s),
\end{equation}
where $\mathcal{S}(x)$ is the feasible solution space defined by problem-specific constraints, and $f(x,s)$ denotes the objective value. A policy network $\pi_{\theta}(\cdot \mid x)$ generates a set of candidate solutions:
\begin{equation}
        \mathcal{S}_x=\{s_i\}_{i=1}^{N} \sim \pi_\theta(\cdot|x).
\end{equation}

For each ordered pair $(s_i,s_j)$, a binary preference label is typically defined as
\begin{equation}
        y_{ij}=\mathbf{1}\{f(x,s_i)<f(x,s_j)\},
\end{equation}
where $y_{ij}=1$ indicates that $s_i$ is preferred over $s_j$. While simple and stable, such binary labels compress the original optimization information: they indicate which solution is better but ignore how much better it is. Consequently, solution pairs with very different quality gaps may receive the same label, weakening the supervision signal and limiting the model’s ability to capture advantage magnitude. This is particularly restrictive in combinatorial optimization, where the magnitude of improvement often reflects meaningful optimization progress. Therefore, preference supervision should move beyond binary comparisons and incorporate both preference direction and advantage magnitude. Another limitation is that existing preference signals are usually applied only after candidate solutions have been generated. The model first constructs a candidate pool and then learns preferences by comparing complete solutions, making preference optimization more like post-hoc re-ranking than search guidance. As a result, performance depends heavily on the quality and diversity of the sampled candidates, while preference information cannot directly influence the construction process. These limitations suggest that preference supervision should be incorporated at multiple levels of constructive search. Local preferences over partial trajectories can guide intermediate decisions toward more promising construction paths, while global preferences over complete solutions ensure consistency with the final optimization objective. Since multi-level preference learning may introduce additional training complexity and computational cost, a flexible training mechanism is also needed to support both training from scratch and efficient adaptation from pretrained checkpoints.

Motivated by these observations, we propose an advantage-aware preference optimization method with two-level constructive search, enabling finer-grained and more search-aware solution optimization.

\section{Methodology}
\begin{figure*}[!t]
	\centering
	\includegraphics[width=1\textwidth]{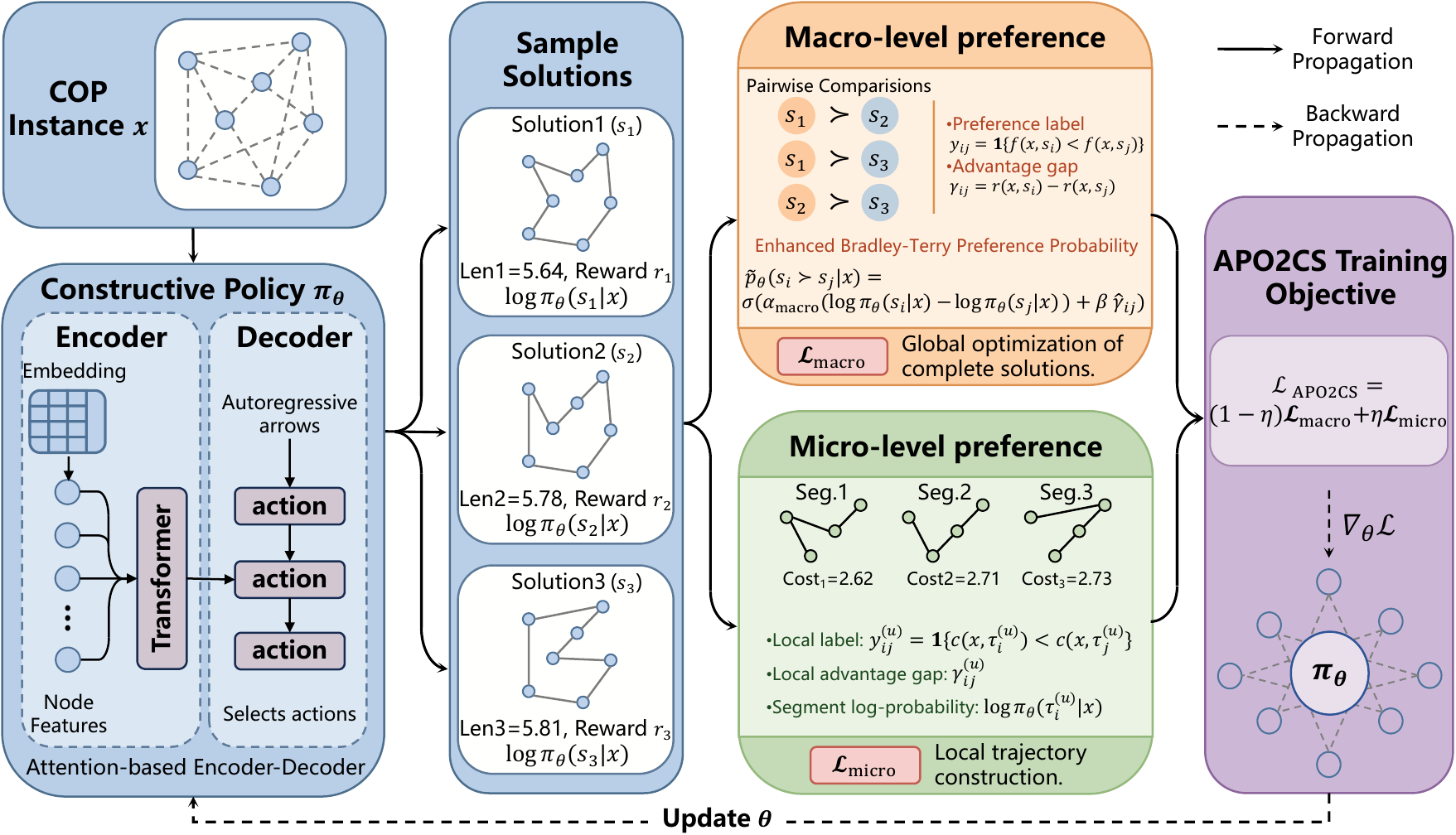}
    \Description{
    Overview of APO2CS. A combinatorial optimization instance is processed
    by a constructive policy to generate candidate solutions and their
    log-probabilities. Complete solutions are used to form advantage-aware
    macro-level preferences, while partial trajectory segments are used to
    form micro-level preferences. The two preference losses are combined to
    optimize the policy through back-propagation.
    }
	\caption{Overview of the APO2CS framework. Given a COP instance $x$, the constructive policy $\pi_\theta$ generates candidate solutions and log-probabilities, from which advantage-aware macro-level preferences over complete solutions and micro-level preferences over trajectory segments are constructed. These two preference losses are integrated into the APO2CS objective to update $\pi_\theta$ via back-propagation.}
	\label{overview}
\end{figure*}
In this section, we present the Advantage-Aware Preference Optimization for Two-Level Constructive Search (APO2CS) framework, as illustrated in Figure~\ref{overview}. APO2CS trains a constructive policy using two complementary preference objectives. At the macro level, it compares complete solutions sampled from the current policy and encourages higher likelihood for solutions with better final objective values, thereby maintaining consistency with the global optimization objective. At the micro level, it decomposes sampled trajectories into local segments and compares partial constructions, providing denser supervision for long sequential rollouts while better respecting local construction constraints. Both objectives adopt advantage-aware pairwise supervision, where pairwise objective differences determine the strength of the preference signal. The final training objective is formulated as a weighted combination of the macro- and micro-level preference losses.

We first introduce the proposed advantage-aware preference optimization, which extends conventional binary preference learning with fine-grained supervision. We then detail the two-level constructive search mechanism that incorporates macro-level and micro-level preferences. APO2CS can be trained either from scratch in an end-to-end manner or efficiently adapted from pretrained checkpoints via lightweight fine-tuning; the details of these two training strategies are provided in Appendix~\ref{End-to-End_Training_and_Lightweight_Fine-Tuning}.

\subsection{Advantage-Aware Preference Optimization}
Preference optimization methods \cite{pbrl} commonly adopt the Bradley–\\Terry model \cite{BT-model} to characterize pairwise preferences between candidate solutions. We first introduce this standard formulation as background before presenting its task-specific instantiation for constructive combinatorial optimization. Specifically, the preference probability that solution $s_i$ is preferred over solution $s_j$ is defined as
\begin{equation}
    p_\theta(s_i \succ s_j \mid x)
    =
    \sigma\!\left(
    \alpha
    \big[
    \log \pi_\theta(s_i \mid x)
    -
    \log \pi_\theta(s_j \mid x)
    \big]
    \right),
\end{equation}
where $s_i$ and $s_j$ are two candidate solutions sampled for instance $x$, $\sigma(\cdot)$ denotes the sigmoid function, and
\(
\alpha[\log \pi_\theta(s_i \mid x)-\log \pi_\theta(s_j \mid x)]
\)
represents the scaled log-likelihood difference between the two solutions.

Maximizing the likelihood over all preference pairs is equivalent to minimizing the following objective:

\begin{equation}
\begin{aligned}
\mathcal{L}_{\mathrm{PO}}(\pi_\theta)
=
-\mathbb{E}_{x \sim \mathcal{X}}
\Bigg[
&\frac{1}{|\mathcal{S}_x|^2}
\sum_{i \neq j} y_{ij} \\
&\cdot
\log \sigma\!\left(
\alpha
\left[
\log \pi_\theta(s_i \mid x)
-
\log \pi_\theta(s_j \mid x)
\right]
\right)
\Bigg].
\end{aligned}
\end{equation}

However, the standard objective only captures the direction of preference through the binary label ($y_{ij}$), while ignoring the magnitude of the quality difference between the two solutions. To preserve such fine-grained information, we define the set of ordered preference pairs for instance $x$ as
\begin{equation}
\mathcal{P}_x^{+}
=
\left\{
(i,j) \mid i \neq j,\;
f(x,s_i) < f(x,s_j)
\right\}.
\end{equation}

where a smaller objective value indicates a better solution. Therefore, for every $(i,j)\in\mathcal{P}_x^{+}$, solution $s_i$ is preferred over solution $s_j$.
For each ordered preference pair, we define the corresponding pairwise quality difference as
\begin{equation}
\gamma_{ij}
=
r(x,s_i)-r(x,s_j),
\qquad
(i,j)\in\mathcal{P}x^{+},
\end{equation}
where $r(x,s)$ denotes the quality score of solution $s$ for instance $x$, defined such that a larger value indicates a better solution. Under this definition, $\gamma{ij}>0$ for every $(i,j)\in\mathcal{P}_x^{+}$.

To make the pairwise differences comparable across instances, we normalize $\gamma_{ij}$ within each candidate-solution set:
\begin{equation}
\hat{\gamma}{ij}
=
\frac{\gamma{ij}}
{
\max\limits_{(p,q)\in\mathcal{P}x^{+}}
\gamma_{pq}
+
\varepsilon
},
\end{equation}
where $\varepsilon$ is a small positive constant introduced for numerical stability. Based on the normalized pairwise difference, we construct the pair-specific coefficient
\begin{equation}
\lambda_{ij}
=
1+\hat{\gamma}_{ij}.
\end{equation}

We further incorporate the normalized pairwise difference into the Bradley--Terry logit, resulting in the enhanced preference score
\begin{equation}
\tilde{p}_\theta(s_i \succ s_j \mid x)
=
\sigma\left(
\alpha
\left[
\log \pi_\theta(s_i\mid x)
-
\log \pi_\theta(s_j\mid x)
\right]
+
\beta\hat{\gamma}_{ij}
\right),
\end{equation}
where $\beta$ is a scaling factor that controls the influence of the pairwise quality difference.

To incorporate the pairwise quality difference into the preference learning objective, we equivalently absorb the coefficient $\lambda_{ij}$ into the preference score as an exponent:
\begin{equation}
\bar{p}_\theta(s_i \succ s_j \mid x)
=
\left[
\tilde{p}_\theta(s_i \succ s_j \mid x)
\right]^{\lambda_{ij}}.
\end{equation}
Here, $\bar{p}_\theta(s_i \succ s_j \mid x)$ denotes a gap-weighted pairwise likelihood score, which adjusts the contribution of each preference pair according to its relative quality difference. Note that it serves as a weighted likelihood term rather than a separately normalized Bradley--Terry probability.

The resulting advantage-aware preference optimization objective is formulated as
\begin{equation}
\begin{aligned}
\mathcal{L}_{\mathrm{APO}}(\pi_\theta)
=
-\mathbb{E}_{x\sim\mathcal{X}}
\Bigg[
\frac{1}{|\mathcal{P}x^{+}|}
\sum_{(i,j)\in\mathcal{P}x^{+}}
\log
\bar{p}_\theta(s_i \succ s_j \mid x)
\Bigg].
\end{aligned}
\end{equation}

\subsection{Theoretical Interpretation}
We show that the proposed advantage-aware preference objective implicitly induces a gap-modulated pairwise policy update.

\paragraph{Proposition 1.}
Let $\pi_\theta$ denote a stochastic policy, and let $\mathcal{P}_x^{+}$ denote the set of ordered preference pairs. We define the advantage-aware preference objective as
\begin{equation}
\begin{aligned}
\mathcal{L}_{\mathrm{APO}}(\theta)
=
- \mathbb{E}_{x\sim\mathcal{X}}
  \Bigg[
  \frac{1}{|\mathcal{P}_x^{+}|}
  \sum_{(i,j)\in\mathcal{P}_x^{+}}
  \log
  \bar{p}_\theta(s_i\succ s_j\mid x)
  \Bigg],
  \end{aligned}
  \end{equation}
  where
\begin{equation}
  \bar{p}_\theta(s_i\succ s_j\mid x)
  =
  \left[
  \sigma\left(
  \alpha
  \left[
  \log \pi_\theta(s_i\mid x)
-
\log \pi_\theta(s_j\mid x)
\right]
+
\beta\hat{\gamma}_{ij}
\right)
\right]^{\lambda_{ij}},
\end{equation}
$\hat{\gamma}_{ij}$ is the normalized pairwise quality gap, and $\lambda_{ij}=1+\hat{\gamma}_{ij}$.

Assuming that $\hat{\gamma}_{ij}$ and $\lambda_{ij}$ are independent of $\theta$, its gradient is
\begin{equation}
\begin{aligned}
\nabla_\theta\mathcal{L}_{\mathrm{APO}}(\theta)
=-&\mathbb{E}_{x\sim\mathcal{X}}
\Bigg[
\frac{1}{|\mathcal{P}_x^{+}|}
\sum_{(i,j)\in\mathcal{P}_x^{+}}
w_{ij}(\hat{\gamma}_{ij})\\
&\cdot
\Big(
\nabla_\theta\log\pi_\theta(s_i\mid x)
-\nabla_\theta\log\pi_\theta(s_j\mid x)
\Big)
\Bigg],
\end{aligned}
\end{equation}
where
\begin{equation}
w_{ij}(\hat{\gamma}_{ij})
=
\alpha\lambda_{ij}
\left[
1-\sigma(z_{ij})
\right],
\end{equation}
and
\begin{equation}
z_{ij}
=
\alpha
\left[
\log\pi_\theta(s_i\mid x)
-\log\pi_\theta(s_j\mid x)
\right]
+
\beta\hat{\gamma}_{ij}.
\end{equation}
Thus, the gradient weight is jointly determined by the normalized quality gap and the current pairwise preference residual. In the small-gap regime, the change in $w_{ij}$ relative to the binary-preference baseline is approximately linear in $\hat{\gamma}_{ij}$.

\paragraph{Proof sketch.}
Since
\begin{equation}
\log\bar{p}_\theta(s_i\succ s_j\mid x)
=\lambda_{ij}\log\sigma(z_{ij}),
\end{equation}
and
\begin{equation}
\nabla_\theta\log\sigma(z_{ij})
=\left[
1-\sigma(z_{ij})
\right]\nabla_\theta z_{ij},
\end{equation}
we have
\begin{equation}
\begin{aligned}
\nabla_\theta z_{ij}
=\alpha
\Big(
\nabla_\theta\log\pi_\theta(s_i\mid x)
-\nabla_\theta\log\pi_\theta(s_j\mid x)
\Big).
\end{aligned}
\end{equation}
Substituting these expressions into the gradient of
$\mathcal{L}_{\mathrm{APO}}$ yields the stated result. Moreover, a first-order Taylor expansion around
$\hat{\gamma}_{ij}=0$ gives
\begin{equation}
w_{ij}(\hat{\gamma}_{ij})
=w_{ij}(0)
+
\mathcal{O}(\hat{\gamma}_{ij}),
\end{equation}

showing that the first-order variation of the gradient weight is linear in the normalized quality gap.

\paragraph{Insight.}
Binary preference optimization learns \emph{which} solution is better, but ignores \emph{how much} better. APO2CS incorporates pairwise quality gaps directly into the objective, yielding gradients modulated by both the improvement magnitude and the current preference residual. This converts preference learning from a pure ranking signal into a \emph{gap-modulated pairwise optimization principle}. Consequently, APO2CS can be viewed as a pairwise policy update that incorporates solution-quality differences without relying on the conventional advantage function used in reinforcement learning.

\subsection{Two-Level Constructive Search}

Although the advantage-aware preference objective improves the ranking consistency among complete solutions, NCO is inherently a sequential decision-making process, where the final solution quality is determined by a series of local choices. Supervising only the complete solution may therefore lead to insufficient guidance for intermediate construction steps. To address this issue, we further propose a Two-Level Constructive Search, which extends advantage-aware preference optimization from the solution level to the partial-construction level. The key idea is to jointly optimize: (i) global preference over complete solutions and (ii) local preference over partial trajectories, so that the policy is encouraged to produce not only better final solutions but also better intermediate decisions during the constructive process.

In NCO model, a candidate solution $s_i=(a_i^1,a_i^2,\dots,a_i^T)$ is generated autoregressively by the constructive policy, where $T$ is the construction horizon and $a_i^t$ denotes the action taken at step $t$. Its likelihood can be factorized as
\begin{equation}
    \pi_\theta(s_i \mid x)
    =
    \prod_{t=1}^{T}
    \pi_\theta(a_i^t \mid x, a_i^{<t}),
\end{equation}
and the corresponding log-likelihood is
\begin{equation}
    \log \pi_\theta(s_i \mid x)
    =
    \sum_{t=1}^{T}
    \log \pi_\theta(a_i^t \mid x, a_i^{<t}).
\end{equation}

\paragraph{Macro-level constructive search.}
At the macro level, we compare complete solutions exactly as in advantage-aware preference optimization. For clarity, we denote the scaling factor in the global Bradley-Terry model by $\alpha_{\mathrm{macro}}$. Then the macro-level preference probability is defined as
\begin{equation}
\begin{aligned}
    \tilde{p}_{\theta}^{\mathrm{macro}}(s_i \succ s_j \mid x)
    =
    &\sigma\!(
    \alpha_{\mathrm{macro}} \\
    &\cdot
    \big[
    \log \pi_\theta(s_i \mid x)
    -
    \log \pi_\theta(s_j \mid x)
    \big]
    +
    \beta \hat{\gamma}_{ij}
    ).
\end{aligned}
\end{equation}
Accordingly, the macro-level loss is written as
\begin{equation}
\begin{aligned}
\mathcal{L}_{\mathrm{macro}}(\pi_\theta)
=
-\mathbb{E}_{x\sim\mathcal{X}}
\Bigg[
\frac{1}{|\mathcal{P}x^{+}|}
\sum_{(i,j)\in\mathcal{P}x^{+}}
\log
\bar{p}_\theta^{\mathrm{macro}}(s_i \succ s_j \mid x)
\Bigg].
\end{aligned}
\end{equation}

This term provides global supervision by encouraging the policy to assign higher probability to better complete solutions.

\paragraph{Micro-level constructive search.}
To provide finer-grained supervision during the construction process, we further divide each solution trajectory into a sequence of local segments. Let $H$ denote a predefined segment length, and let $\tau_i^{(u)}$ be the $u$-th segment extracted from solution $s_i$, where $u=1,\dots,U$ and $U$ is the number of extracted segments. The likelihood of a segment is defined as the product of step-wise action probabilities within the corresponding interval:
\begin{equation}
    \pi_\theta(\tau_i^{(u)} \mid x)
    =
    \prod_{t \in \mathcal{I}_u}
    \pi_\theta(a_i^t \mid x, a_i^{<t}),
\end{equation}
where $\mathcal{I}_u$ denotes the set of time indices belonging to the $u$-th segment. Its log-likelihood is thus
\begin{equation}
    \log \pi_\theta(\tau_i^{(u)} \mid x)
    =
    \sum_{t \in \mathcal{I}_u}
    \log \pi_\theta(a_i^t \mid x, a_i^{<t}).
\end{equation}

For each segment, we compute a local segment cost $c(x,\tau_i^{(u)})$, which evaluates the partial construction quality within that segment. Since lower cost indicates a better partial trajectory, the local preference label is defined as
\begin{equation}
    y_{ij}^{(u)}
    =
    \mathbf{1}
    \left\{
    c(x,\tau_i^{(u)}) < c(x,\tau_j^{(u)})
    \right\}.
\end{equation}
Correspondingly, the local quality gap is defined as
\begin{equation}
    \gamma_{ij}^{(u)}
    =
    r^{(u)}(x,\tau_i^{(u)})
    -
    r^{(u)}(x,\tau_j^{(u)}),
\end{equation}
where $r^{(u)}(x,\tau)=-c(x,\tau)$ denotes the segment-quality score. Similar to the global case, we introduce a normalized local weight
\begin{equation}
    \hat{\gamma}_{ij}^{(u)}
    =
    1+
    \frac{\gamma_{ij}^{(u)}}{
    \max\limits_{(p,q)\in \mathcal{P}_{x,u}^{+}}
    \gamma_{pq}^{(u)}
    +\varepsilon},
\end{equation}
where$\mathcal{P}_{x,u}^{+}
=\Big\{(i,j)|i\neq j,r^{(u)}(x,\tau_i^{(u)})>r^{(u)}(x,\tau_j^{(u)})\Big\}$ denotes the ordered local preference-pair set. Based on these local quantities, we define the micro-level preference probability by an advantage-enhanced Bradley-Terry model:
\begin{equation}
\begin{aligned}
&\tilde{p}_{\theta}^{\mathrm{micro}}
\bigl(\tau_i^{(u)} \succ \tau_j^{(u)} \mid x\bigr) \\
&\quad =
\sigma\Bigl(
\alpha_{\mathrm{micro}}
\bigl[
\log \pi_\theta(\tau_i^{(u)} \mid x)
-
\log \pi_\theta(\tau_j^{(u)} \mid x)
\bigr]
+
\beta \hat{\gamma}_{ij}^{(u)}
\Bigr).
\end{aligned}
\end{equation}
where $\alpha_{\mathrm{micro}}$ controls the sensitivity to segment-level likelihood differences. The micro-level loss is then defined as
\begin{equation}
\begin{aligned}
\mathcal{L}_{\mathrm{micro}}(\pi_\theta)
=
-\mathbb{E}_{x \sim \mathcal{X}}
\Bigg[
&\frac{1}{U}
\sum_{u=1}^{U}
\frac{1}{|\mathcal{P}_{x,u}^{+}|}
\sum_{(i,j)\in\mathcal{P}_{x,u}^{+}}\\
&\qquad
\log
\bar{p}_{\theta}^{\mathrm{micro}}
\left(
\tau_i^{(u)}
\succ
\tau_j^{(u)}
\mid x
\right)
\Bigg].
\end{aligned}
\end{equation}
This objective explicitly regularizes the local construction behavior by encouraging the policy to assign higher probability to better partial trajectories.

\paragraph{Joint optimization.}
Finally, the two levels are integrated into a unified training objective:
\begin{equation}
    \mathcal{L}_{\mathrm{APO2CS}}(\pi_\theta)
    =
    (1-\eta)\,
    \mathcal{L}_{\mathrm{macro}}(\pi_\theta)
    +
    \eta\,
    \mathcal{L}_{\mathrm{micro}}(\pi_\theta),
\end{equation}
where $\eta \in [0,1]$ controls the trade-off between global solution ranking and local trajectory refinement. When $\eta=0$, the objective reduces to purely global preference optimization; when $\eta>0$, the policy is additionally guided by fine-grained local preferences.

The two objectives provide complementary guidance. The macro-level objective preserves the global search direction by comparing complete solutions according to their final quality, whereas the micro-level objective alleviates long-horizon credit assignment by introducing preference signals into intermediate construction steps. Their combination allows the policy to learn both which complete solutions are preferable and which local decisions contribute to constructing them.

\section{Experiments}
In this section, we present the main experimental results to evaluate the effectiveness of the proposed APO2CS algorithm for COPs. Specifically, we aim to answer the following research questions:
\begin{itemize}
 \item \textbf{RQ1.} How does APO2CS perform in comparison with existing algorithms on standard benchmarks, such as the Traveling Salesman Problem (TSP), the Capacitated Vehicle Routing Problem (CVRP) and the knapsack problem (KP)? 
 \item \textbf{RQ2.} How effective is the proposed two-level preference optimization mechanism, compared with conventional preference optimization methods, for constructive search? 
 \item \textbf{RQ3.} How flexible and efficient is the proposed training paradigm under different initialization settings, including cold-start and warm-start training?
\end{itemize}
RQ1 is addressed through the main results presented in Section~\ref{main_results}, where APO2CS is compared with representative baselines. RQ2 and RQ3 are investigated in Sections~\ref{Ablation_Studies} and~\ref{Analysis_of_Startup_Strategy}, respectively, through comprehensive ablation studies and startup strategy analyses.

\subsection{Experimental Setup}
\paragraph{Benchmark.} APO2CS is a reinforcement learning based neural solver designed for COPs. To evaluate its effectiveness, we consider three fundamental COP benchmarks as our testbeds: the TSP, the CVRP, and the KP. In the main experiments, for TSP and CVRP, we follow Kool et al. \cite{kool2019attentionlearnsolverouting} and generate problem instances by uniformly sampling node coordinates. For KP, we follow the setup by Bello et al. \cite{bello2017neural}. For each routing problem, we evaluate three standard scales with 20, 50, and 100 nodes. For KP, we evaluate three problem scales with 50, 100, and 200 items. In TSP, the objective is to identify a minimum-length Hamiltonian tour over all nodes in the graph. In CVRP, the objective is to design a set of routes that start from and return to the depot, while satisfying vehicle capacity constraints and customer demand requirements, so as to minimize the total travel distance. For both routing problems, the reward is defined as the Euclidean length of the resulting route. In KP, each instance consists of a set of items with associated weights and values, and the objective is to select a subset of items that maximizes the total value while satisfying the knapsack capacity constraint. Accordingly, the reward for KP is defined as the total value of the selected items. All experiments are performed on a server with NVIDIA RTX A6000 GPUs (48 GB) and an Intel(R) Xeon(R) Gold 6230 CPU @ 2.10 GHz.

\paragraph{Baselines.} We assess the optimality gap using strong problem-specific solvers. For TSP and CVRP, we use Concorde \cite{concorde} and LKH3 \cite{LKH3}, respectively, as widely adopted heuristic solvers. For KP, we compute the optimal solution using dynamic programming and report the optimality gap with respect to the DP results. In addition, we include a greedy heuristic baseline for KP, which selects items according to a hand-crafted value-to-weight criterion. We further compare APO2CS against reinforcement-learning-based neural solvers trained with variants of REINFORCE, including Pointer Net \cite{bello2017neural}, AM \cite{kool2019attentionlearnsolverouting}, and POMO \cite{pomo}. AM estimates the advantage function based on the solution generated at the previous step, while POMO refines this estimate by averaging the quality of multiple sampled solutions from the same instance and using them as baselines. For KP, we also compare with MTL-COP \cite{mtlcop}, a multi-task learning based neural solver designed for combinatorial optimization problems. We additionally include comparisons with preference-optimization-based neural solvers, namely BOPO \cite{bopo} and POCO \cite{poco}, both of which are built upon the POMO framework. BOPO enhances sample efficiency through three specifically designed components, whereas POCO improves the trade-off between exploitation and exploration by introducing a reparameterized entropy-regularized objective. Unless otherwise stated, all hyperparameter settings, network architectures, and optimizer configurations are kept consistent with those in the original papers.

\paragraph{Evaluation.} Inference follows a unified protocol: 1) Each validation set consists of exactly 10,000 instances. 2) We apply 8-fold augmentation (denoted as x8 aug.) to each instance. 3) The metrics include the average objective value of the solutions on the validation set, the average optimality gap, and the total time to solve these instances.

\subsection{Main Results (RQ1)}
\label{main_results}
The performance of APO2CS on TSP, CVRP, and KP is reported in Table \ref{main_tsp} and Table \ref{main_kp}. Overall, APO2CS achieves the best or near-best results among neural baselines across problem scales, demonstrating strong optimization capability in constructive search. Results on KP further confirm its flexibility in handling different types of combinatorial optimization problems.

Without augmentation, APO2CS attains the lowest optimality gaps among neural solvers on all TSP and CVRP benchmark settings. Its advantage is particularly clear on larger instances, such as TSP-100 and CVRP-100, where it outperforms both RL-based baselines and existing preference-optimization-based methods. With x8 augmentation, APO2CS further improves performance on most routing settings and remains the strongest neural method overall. On TSP, the augmented version achieves an almost zero gap and stays very close to LKH3. On CVRP, augmentation brings additional gains on small and medium instances, while yielding only marginal improvements on larger ones. Moreover, APO2CS offers an attractive trade-off between solution quality and efficiency. Although strong heuristic solvers such as Concorde and LKH3 remain important references, they require substantially higher runtime, whereas APO2CS achieves near-heuristic solution quality with much lower inference cost.

To further evaluate its flexibility beyond routing problems, we conduct additional experiments on KP. We adapt APO2CS by representing each item with its weight and value, replacing the $x$- and $y$-coordinates used for TSP nodes. During decoding, the generated item sequence is inserted into the knapsack sequentially until the capacity is reached, after which decoding terminates. As shown in Table \ref{main_kp}, APO2CS achieves the best or tied-best performance among neural baselines on KP50, KP100, and KP200. Compared with greedy heuristics, Pointer Net, AM, POMO, and MTL-COP, it consistently obtains smaller optimality gaps relative to the optimal DP solutions, reaching only 0.02\%, 0.01\%, and 0.01\% on KP50, KP100, and KP200, respectively. These results indicate that APO2CS can effectively adapt to non-routing COPs, where the objective is subset selection rather than route construction.

In summary, Table \ref{main_tsp} and Table \ref{main_kp} show that APO2CS consistently outperforms existing neural baselines on TSP and CVRP and generalizes well to KP. These results highlight its effectiveness and flexibility across different problem sizes and structures, while maintaining competitive performance against strong problem-specific solvers.

\begin{table*}[!t]
  \centering
  \caption{Experiment results on TSP and CVRP. Gap is evaluated on 10k instances, and Time is the sum of them.}
            \resizebox{\textwidth}{!}{
    \begin{tabular}{cccccccccc|ccccccccc}
    \toprule
    \multirow{3}[6]{*}{\textbf{Method}} & \multicolumn{9}{c|}{\textbf{TSP}}                                     & \multicolumn{9}{c}{\textbf{CVRP}} \\
\cmidrule{2-19}          & \multicolumn{3}{c}{\textbf{N=20}} & \multicolumn{3}{c}{\textbf{N=50}} & \multicolumn{3}{c|}{\textbf{N=100}} & \multicolumn{3}{c}{\textbf{N=20}} & \multicolumn{3}{c}{\textbf{N=50}} & \multicolumn{3}{c}{\textbf{N=100}} \\
\cmidrule(lr){2-4}\cmidrule(lr){5-7}\cmidrule(lr){8-10}\cmidrule(lr){11-13}\cmidrule(lr){14-16}\cmidrule(lr){17-19}          & Obj. & Gap\% & Time  & Obj. & Gap\% & Time  & Obj. & Gap\% & Time  & Obj. & Gap\% & Time  & Obj. & Gap\% & Time  & Obj. & Gap\% & Time \\
    \midrule
    Concorde & 3.833 & -     & 5m    & 5.690  & -     & 13m   & 7.761 & -     & 1h    & -     & -     & -     & -     & -     & -     & -     & -     & - \\
    LKH3  & 3.832 & 0.00  & 42s   & 5.693 & 0.00  & 6m    & 7.762 & 0.00  & 25m   & 6.120  & 0.00  & 2h    & 10.409 & 0.00  & 7h    & 15.687 & 0.00  & 12h \\
    \midrule
    AM    & 3.846 & 0.37  & 1s    & 5.779 & 1.51  & 1s    & 8.039 & 3.57  & 2s    & 6.401 & 4.59  & 1s    & 10.934 & 5.04  & 1s    & 16.781 & 6.97  & 3s \\
    POMO  & 3.835 & 0.08  & 1s    & 5.700   & 0.12  & 2s    & 7.795 & 0.43  & 11s   & 6.175 & 0.90  & 1s    & 10.496 & 0.84  & 4s    & 15.832 & 0.92  & 19s \\
    BOPO  & 3.839 & 0.08  & 3s    & 5.702 & 0.12  & 6s    & 7.787 & 0.32  & 12s   & 6.181 & 1.00  & 1s    & 10.502 & 0.89  & 5s    & 15.843 & 0.99  & 20s \\
    POCO  & 3.841 & 0.23  & 3s    & 5.701 & 0.14  & 6s    & 7.775 & 0.17  & 18s   & 6.175 & 0.90   & 1s    & 10.496 & 0.84  & 6s    & 15.794 & 0.68  & 21s \\
    \rowcolor[rgb]{ .949,  .949,  .949} \textbf{APO2CS} & \cellcolor[rgb]{ 1,  1,  1} \textbf{3.832 } & \cellcolor[rgb]{ 1,  1,  1} \textbf{0.00 } & \cellcolor[rgb]{ 1,  1,  1} 2s & \cellcolor[rgb]{ 1,  1,  1} \textbf{5.699} & \cellcolor[rgb]{ 1,  1,  1} \textbf{0.11} & \cellcolor[rgb]{ 1,  1,  1} 5s & \cellcolor[rgb]{ 1,  1,  1} \textbf{7.770 } & \cellcolor[rgb]{ 1,  1,  1} \textbf{0.10 } & \cellcolor[rgb]{ 1,  1,  1} 13s & \cellcolor[rgb]{ 1,  1,  1} \textbf{6.159} & \cellcolor[rgb]{ 1,  1,  1} \textbf{0.64} & \cellcolor[rgb]{ 1,  1,  1} 1s & \cellcolor[rgb]{ 1,  1,  1} \textbf{10.425} & \cellcolor[rgb]{ 1,  1,  1} \textbf{0.15} & \cellcolor[rgb]{ 1,  1,  1} 4s & \cellcolor[rgb]{ 1,  1,  1} \textbf{15.725} & \cellcolor[rgb]{ 1,  1,  1} \textbf{0.24} & \cellcolor[rgb]{ 1,  1,  1} 18s \\
    \midrule
    POMO (x8 aug.) & 3.834 & 0.05  & 3s    & 5.695 & 0.04  & 16s   & 7.772 & 0.13  & 1m    & 6.142 & 0.36  & 5s    & 10.426 & 0.16  & 26s   & 15.731 & 0.28  & 2m \\
    BOPO (x8 aug.) & 3.835 & 0.08  & 3s    & 5.697 & 0.07  & 6s    & 7.779 & 0.22  & 18s   & 6.143 & 0.38  & 4s    & 10.429 & 0.19  & 19s   & 15.737 & 0.32  & 29s \\
    POCO (x8 aug.) & 3.839 & 0.18  & 6s    & 5.694 & 0.02  & 17s   & 7.774 & 0.15  & 1m    & 6.139 & 0.31  & 5s    & 10.418 & 0.09  & 21s   & 15.731 & 0.28  & 2m \\
    \rowcolor[rgb]{ .949,  .949,  .949} \textbf{APO2CS (x8 aug.)} & \cellcolor[rgb]{ 1,  1,  1} \textbf{3.832} & \cellcolor[rgb]{ 1,  1,  1} \textbf{0.00 } & \cellcolor[rgb]{ 1,  1,  1} 6s & \cellcolor[rgb]{ 1,  1,  1} \textbf{5.693} & \cellcolor[rgb]{ 1,  1,  1} \textbf{0.00 } & \cellcolor[rgb]{ 1,  1,  1} 17s & \cellcolor[rgb]{ 1,  1,  1} \textbf{7.763} & \cellcolor[rgb]{ 1,  1,  1} \textbf{0.01} & \cellcolor[rgb]{ 1,  1,  1} 1m & \cellcolor[rgb]{ 1,  1,  1} \textbf{6.132} & \cellcolor[rgb]{ 1,  1,  1} \textbf{0.20 } & \cellcolor[rgb]{ 1,  1,  1} 5s & \cellcolor[rgb]{ 1,  1,  1} \textbf{10.417} & \cellcolor[rgb]{ 1,  1,  1} \textbf{0.08} & \cellcolor[rgb]{ 1,  1,  1} 20s & \cellcolor[rgb]{ 1,  1,  1} \textbf{15.726} & \cellcolor[rgb]{ 1,  1,  1} \textbf{0.25} & \cellcolor[rgb]{ 1,  1,  1} 2m \\
    \bottomrule
    \end{tabular}%
    }
  \label{main_tsp}%
\end{table*}%

\begin{table*}[!t]
  \centering
  \caption{Experiment results on KP. Gap is evaluated on 10k instances.}
            \resizebox{0.6\textwidth}{!}{
    \begin{tabular}{ccccccc}
    \toprule
    \multirow{2}[4]{*}{\textbf{Method}} & \multicolumn{2}{c}{\textbf{KP50}} & \multicolumn{2}{c}{\textbf{KP100}} & \multicolumn{2}{c}{\textbf{KP200}} \\
\cmidrule(lr){2-3}\cmidrule(lr){4-5}\cmidrule(lr){6-7}          & Score & Gap\%   & Score & Gap\%   & Score & Gap\% \\
    \midrule
    Optimal / DP & 20.128  & 0.00     & 40.458  & 0.00     & 57.602  & 0.00 \\
    Greedy Heuristics & 19.907  & 1.10 & 40.205  & 0.63 & 57.247  & 0.62 \\
    Pointer Net & 19.914  & 1.06 & 40.218  & 0.59 & 57.262  & 0.59 \\
    AM    & 19.961  & 0.83 & 40.250  & 0.51 & 57.278  & 0.56 \\
    POMO  & 20.120  & 0.04 & 40.452  & 0.01 & 57.594  & 0.01 \\
    MTL-COP & 20.122  & 0.03 & 40.452  & 0.01 & 57.599  & 0.01 \\
    \rowcolor[rgb]{ .949,  .949,  .949} \textbf{APO2CS} & \cellcolor[rgb]{ 1,  1,  1} \textbf{20.123 } & \cellcolor[rgb]{ 1,  1,  1} \textbf{0.02} & \cellcolor[rgb]{ 1,  1,  1} \textbf{40.453 } & \cellcolor[rgb]{ 1,  1,  1} \textbf{0.01} & \cellcolor[rgb]{ 1,  1,  1} \textbf{57.599 } & \cellcolor[rgb]{ 1,  1,  1} \textbf{0.01} \\
    \bottomrule
    \end{tabular}%
    }
  \label{main_kp}%
\end{table*}%

\begin{figure*}[t]
    \centering

    \begin{subfigure}{0.32\textwidth}
        \centering
        \includegraphics[width=\linewidth]{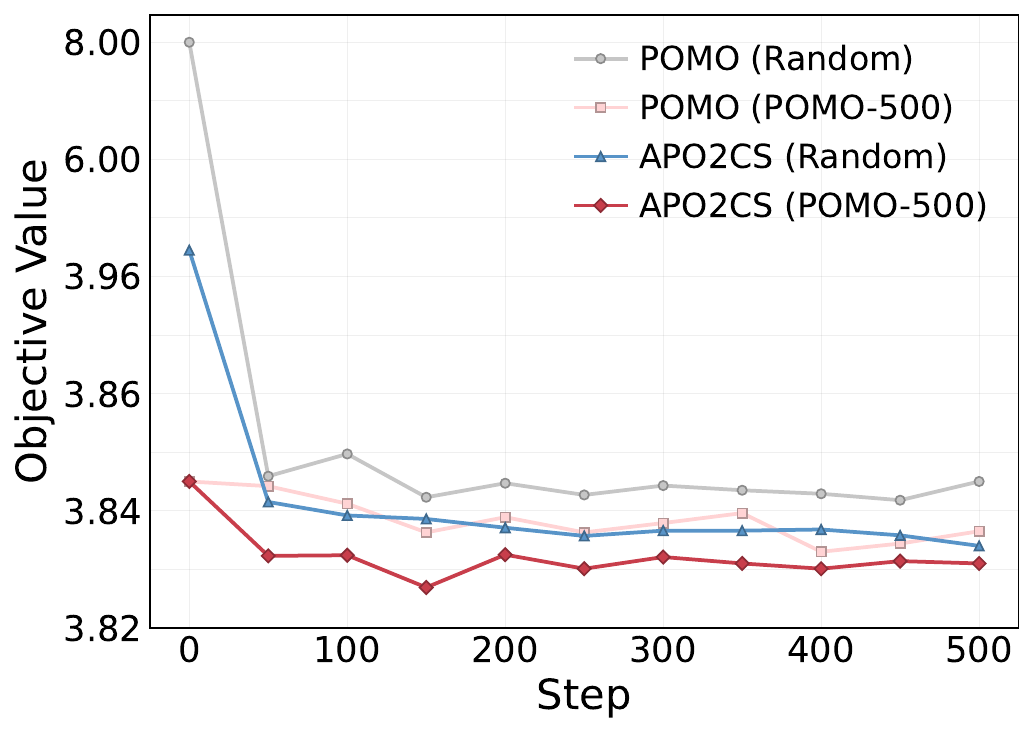}
        \caption{N=20}
        \label{fig:tsp20}
    \end{subfigure}
    \hfill
    \begin{subfigure}{0.32\textwidth}
        \centering
        \includegraphics[width=\linewidth]{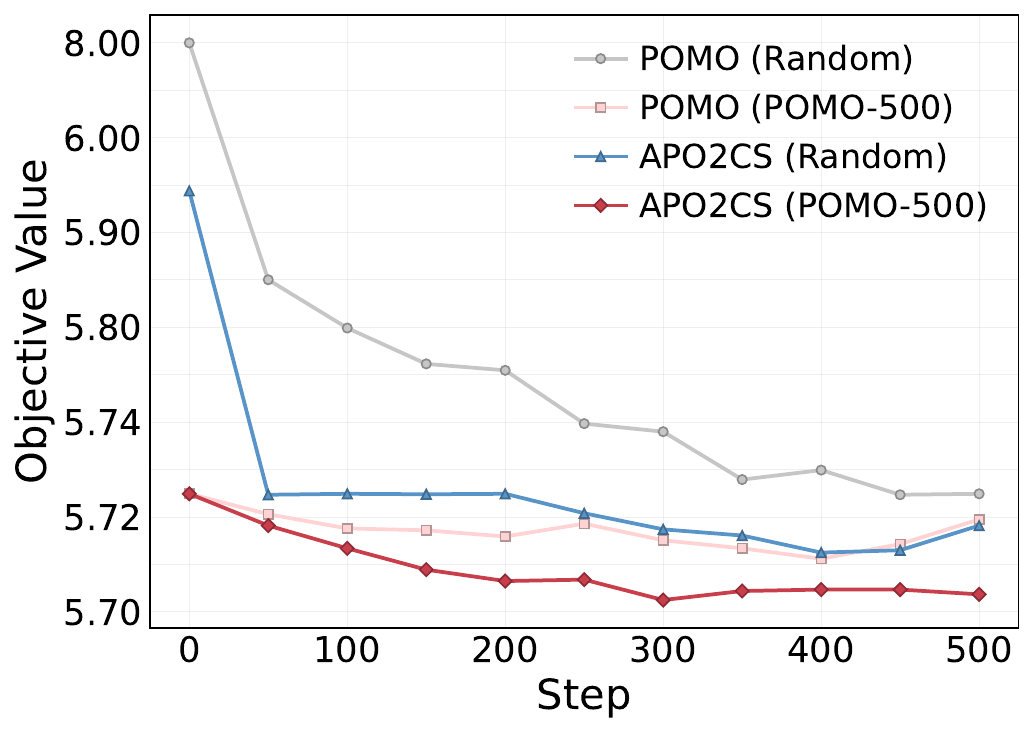}
        \caption{N=50}
        \label{fig:tsp50}
    \end{subfigure}
    \hfill
    \begin{subfigure}{0.32\textwidth}
        \centering
        \includegraphics[width=\linewidth]{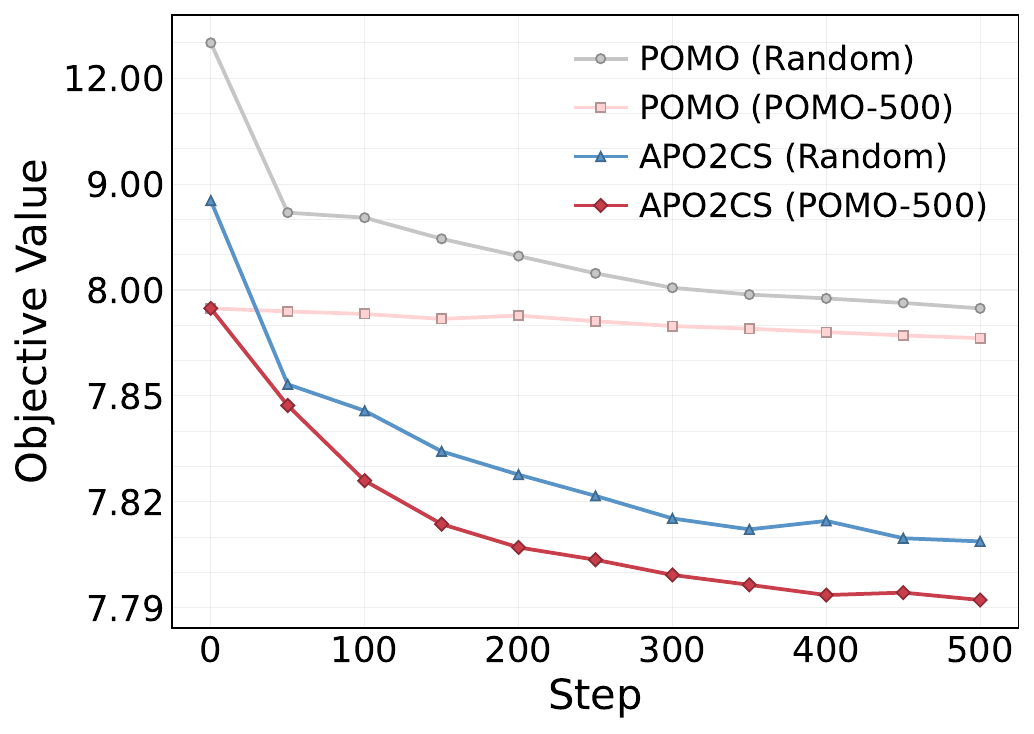}
        \caption{N=100}
        \label{fig:tsp100}
    \end{subfigure}

    \Description{
    Three subfigures compare POMO and APO2CS on TSP instances with different numbers of nodes. The plots show performance comparisons for instances with 20, 50, and 100 nodes under random initialization and POMO-based warm-start initialization.}

    \caption{Performance comparison of POMO and APO2CS on TSP instances with different sizes. "Random" denotes training from scratch, while "POMO-500" denotes warm-start initialization from a pretrained POMO model after 500 epochs.}
    \label{fig:three_subfigures}
\end{figure*}

\subsection{Ablation Studies (RQ2)}
\label{Ablation_Studies}
We conduct systematic ablation studies to evaluate the contributions of advantage-aware preference optimization and the two-level constructive search mechanism. Specifically, we compare four variants: POMO*, which augments POMO with conventional binary-label preference optimization; APO2CS\_macro, which retains only the macro-level preference loss; APO2CS\_micro, which retains only the micro-level preference loss; and the full APO2CS model.
\begin{table}[t]
  \centering
  \caption{Ablation experimental results on TSP. Gap is evaluated on 10k instances, and Time is the sum of them.}
            \resizebox{\columnwidth}{!}{
    \begin{tabular}{ccccccc}
    \toprule
    \multirow{2}[4]{*}{\textbf{Method}} & \multicolumn{3}{c}{\textbf{N=50}} & \multicolumn{3}{c}{\textbf{N=100}} \\
\cmidrule(lr){2-4}\cmidrule(lr){5-7}          & Obj.  & Gap\% & Time  & Obj.  & Gap\% & Time \\
    \midrule
    POMO* & 5.722 & 0.51  & 6s    & 7.808 & 0.59  & 18s \\
    APO2CS\_macro & 5.712 & 0.33  & 5s    & 7.79  & 0.36  & 12s \\
    APO2CS\_micro & 7.035 & 23.57  & 6s    & 10.052 & 29.50  & 14s \\
    APO2CS & \textbf{5.701} & \textbf{0.14 } & 5s    & \textbf{7.77} & \textbf{0.10 } & 13s \\
    \bottomrule
    \end{tabular}%
    }
  \label{Ablation}%
\end{table}%

Table~\ref{Ablation} reports the ablation results on TSP with problem sizes N=50 and N=100. The comparison of APO2CS\_macro, APO2CS\_micro, and APO2CS clearly shows that all components of APO2CS are necessary. Using only the micro-level preference loss leads to severe performance degradation, yielding gaps of 23.57\% and 29.50\%, respectively. This indicates that local trajectory-level supervision alone is insufficient to ensure high global solution quality, as micro-level preference primarily captures short-horizon construction quality and may bias the policy toward locally favorable yet globally suboptimal decisions. In contrast, using only the macro-level preference loss already achieves much stronger performance, with gaps of 0.33\% and 0.36\%, confirming the importance of global solution-level preference learning. Nevertheless, the full APO2CS further reduces the gap to 0.14\% and 0.10\%, respectively, demonstrating that micro-level preference provides useful complementary guidance when combined with macro-level supervision. Specifically, the macro-level aligns optimization with final solution quality, whereas the micro-level supplies denser local supervision during constructive search.

Moreover, the comparison between POMO* and APO2CS\_macro highlights the advantage of the proposed advantage-aware preference optimization over conventional binary preference optimization. Although both variants rely solely on global solution-level preference supervision, APO2CS\_macro consistently outperforms POMO*, reducing the gap from 0.51\% to 0.33\% on N=50 and from 0.59\% to 0.36\% on N=100. This improvement suggests that incorporating advantage magnitude into the preference signal provides more informative supervision than binary labels alone. Rather than learning only which solution is better, the model can additionally exploit how much better one solution is than another, leading to stronger preference discrimination and more effective optimization.

\subsection{Analysis of Startup Strategy (RQ3)}
\label{Analysis_of_Startup_Strategy}

We evaluate the effect of different initialization strategies in APO2CS by comparing cold-start training from random initialization with warm-start training from a pretrained POMO-500 checkpoint. The results shown in Figure \ref{fig:three_subfigures} demonstrate that APO2CS already outperforms POMO under the cold-start setting, indicating the effectiveness of the proposed method itself. However, cold-start training begins from a relatively poor policy and therefore requires more optimization steps to reach stable performance. By contrast, APO2CS with warm-start consistently achieves lower objective values across TSP-N20, TSP-N50, and TSP-N100. Since warm-start uses the pretrained POMO-500 model as initialization, it avoids the inefficient exploration stage in early training and starts from a much stronger policy. More importantly, when compared with continued POMO training from the same POMO-500 checkpoint, APO2CS still obtains better results, showing that the gain is not only due to improved initialization, but also to the effectiveness of the proposed APO2CS optimization mechanism. In addition, warm-start reaches better performance with fewer additional training steps, demonstrating higher training efficiency and better utilization of the available training budget. Overall, these results verify the standalone effectiveness of APO2CS under cold-start training and further confirm that warm-start serves as a complementary strategy for accelerating convergence and improving final solution quality.

\section{Conclusion}
In this paper, we proposed APO2CS, an advantage-aware preference optimization framework for two-level constructive search in combinatorial optimization. Unlike conventional methods based on binary preference labels, APO2CS captures both preference direction and advantage magnitude, allowing the model to better distinguish solution-quality differences. It further combines macro-level preferences over complete solutions with micro-level preferences over partial construction trajectories, enabling preference signals to guide both global optimization and local decision making during generation. APO2CS also supports both cold-start training and warm-start fine-tuning, making it flexible under different computational budgets. Experiments on TSP and CVRP show that APO2CS consistently achieves strong performance across problem scales, outperforming existing reinforcement-learning-based and preference-optimization-based neural solvers while maintaining competitive inference efficiency. In addition, we further evaluate APO2CS on KP to examine its flexibility beyond routing problems. The results on KP demonstrate that APO2CS achieves highly competitive performance against both heuristic and neural baselines, indicating its potential applicability to diverse combinatorial optimization structures beyond routing.

Despite these encouraging results, APO2CS still has a limitation: its micro-level preference pairs rely on predefined trajectory segmentation, which may not always match the most informative decision boundaries across different problem domains. Future work will explore adaptive mechanisms for constructing local preference signals and further investigate the theoretical properties of advantage-aware preference optimization in constructive search. We also plan to extend APO2CS to broader constrained and heterogeneous combinatorial optimization problems, particularly those arising in parallel and distributed training.

\section*{Ethics and Privacy Statement}
This work focuses on developing learning-based optimization algorithms for combinatorial optimization problems. It does not involve human subjects, personal data collection, or sensitive information. The proposed method is intended for algorithmic decision-making and does not introduce additional privacy risks beyond those associated with existing optimization approaches.

\balance
\bibliographystyle{ACM-Reference-Format}
\bibliography{sample-base}

\newpage
\appendix

\section{Appendix}
\subsection{End-to-End Training and Lightweight Fine-Tuning}
\label{End-to-End_Training_and_Lightweight_Fine-Tuning}
To make APO2CS more practical under different computational budgets, we introduce a flexible training strategy that supports both \emph{cold-start} and \emph{warm-start} initialization. APO2CS can be trained either from scratch or from a pretrained checkpoint, reducing training cost and computational overhead while maintaining the benefits of advantage-aware two-level preference optimization.

In the cold-start setting, the policy network is randomly initialized and trained end-to-end on the target combinatorial optimization task. For each mini-batch, the model samples candidate solutions, builds macro- and micro-level preference pairs, computes the APO2CS loss, and updates the parameters via backpropagation. This setting requires no external initialization and enables the model to learn both solution construction and preference-guided optimization directly from data. When a pretrained constructive policy is available, APO2CS can be initialized from the checkpoint and further fine-tuned with the same preference-based objective. The pretrained model offers a strong starting point for solution generation, while macro- and micro-level preference supervision further improves its search behavior. Compared with training from scratch, warm-start fine-tuning is more efficient and requires fewer optimization steps. The overall training process is summarized in Algorithm~\ref{alg:training}.

\begin{algorithm}[t]
\caption{APO2CS Training Protocol}
\label{alg:training}
\begin{algorithmic}[1]
\REQUIRE Epochs $E$, dataset $\mathcal{D}$, policy $\pi_\theta$, mode $m \in \{\text{cold}, \text{warm}\}$, optional checkpoint $\theta_0$, segment level $u$, candidate number $N$, learning rate $\alpha$.
\IF{$m=\text{warm}$}
    \STATE Initialize $\theta \leftarrow \theta_0$
\ELSE
    \STATE Initialize $\theta$ randomly
\ENDIF
\FOR{epoch $e=1$ to $E$}
    \FOR{each mini-batch $\mathcal{B} \subset \mathcal{D}$}
        \FOR{each $x \in \mathcal{B}$}
            \STATE Sample candidates $\mathcal{S}_x=\{\tau_i\}_{i=1}^{N} \sim \pi_\theta(\cdot \mid x)$
            \STATE Extract level-$u$ segments $\mathcal{S}_x^{(u)}$
            \STATE Construct macro-level pairs $\mathcal{P}_x$ from $\mathcal{S}_x$
            \STATE Construct micro-level pairs $\mathcal{P}_x^{(u)}$ from $\mathcal{S}_x^{(u)}$
        \ENDFOR
        \STATE Compute $\mathcal{L}_{\mathrm{macro}}$ and $\mathcal{L}_{\mathrm{micro}}$
        \STATE $\mathcal{L}_{\mathrm{APO2CS}} \leftarrow (1-\eta)\mathcal{L}_{\mathrm{macro}} + \eta\mathcal{L}_{\mathrm{micro}}$
        \STATE Update $\theta \leftarrow \theta - \alpha \nabla_\theta \mathcal{L}_{\mathrm{APO2CS}}$
    \ENDFOR
\ENDFOR
\end{algorithmic}
\end{algorithm}

\subsection{Train Details}

Table~\ref{tab:apo2cs-general-train-params} summarizes the general training hyperparameters used for APO2CS. We evaluate APO2CS on three problem families, including TSP, CVRP, and KP. For TSP and CVRP, the problem sizes are set to 20, 50, and 100, while for KP, the problem sizes are set to 50, 100, and 200.
\begin{table}[t]
\centering
\caption{General training hyperparameters for APO2CS.}
\label{tab:apo2cs-general-train-params}
\setlength{\tabcolsep}{5pt}
\renewcommand{\arraystretch}{1.12}
\resizebox{0.85\linewidth}{!}{
\begin{tabular}{ccc}
\toprule
\textbf{Category} & \textbf{Parameter} & \textbf{Value / Setting} \\
\midrule
\multirow{1}{*}{\textbf{Problem setting}}
& Problem sizes
& \begin{tabular}[c]{@{}c@{}}
20, 50, 100 for TSP and CVRP \\
50, 100, 200 for KP
\end{tabular} \\
\midrule
\multirow{7}{*}{\textbf{Model architecture}}
& Embedding dimension & 128 \\
& Encoder layers & 6 \\
& Decoder layers & 1 \\
& QKV dimension & 16 \\
& Attention heads & 8 \\
& Feed-forward hidden dimension & 512 \\
& Multi-start hidden dimension & 16 \\
\midrule
\multirow{8}{*}{\textbf{Optimization}}
& Optimizer & Adam \\
& Learning rate & $3\times10^{-4}$ \\
& Weight decay & $1\times10^{-6}$ \\
& Epochs & 2000 \\
& Training episodes & 50000 \\
& Batch size & 64 \\
& Optimizer step interval & 1 \\
& Data augmentation & Disabled \\
\midrule
\multirow{3}{*}{\textbf{Preference optimization}}
& Loss type & \texttt{dual\_po\_loss} \\
& $\eta$ & 0.5 \\
& Segment length & 5 \\
\bottomrule
\end{tabular}
}
\end{table}

For the model architecture, we use a unified configuration across tasks. The model adopts a 128-dimensional embedding space, 6 encoder layers, 1 decoder layer, 8 attention heads, 16-dimensional QKV projections, and a 512-dimensional feed-forward hidden layer. The multi-start hidden dimension is set to 16.

For optimization, we use the Adam optimizer with a learning rate of $3\times10^{-4}$ and a weight decay of $1\times10^{-6}$. The model is trained for 2000 epochs with 50000 training episodes and a batch size of 64. The optimizer is updated every step, and data augmentation is disabled during training.

For preference optimization, we use \texttt{dual\_po\_loss} as the training objective. The balancing coefficient $\eta$ is set to 0.5, and the segment length for local preference modeling is fixed to 5 across all experiments.

\subsection{Licenses for existing assets}
The assets used in this work are listed in Table \ref{assets} and are all open-source for academic research. We will release our source code under the MIT License.

\begin{table}[t]
  \centering
  \caption{Assets, Licenses and Their Usage}
  \resizebox{0.85\linewidth}{!}{
    \begin{tabular}{cccc}
    \toprule
    \textbf{Type} & \textbf{Asset} & \textbf{License} & \textbf{Usage} \\
    \midrule
    \multirow{8}[2]{*}{\textbf{Code}} & Concorde(David et al. 2011) & Available for academic use & Evaluation \\
          & Pointer Net(Bello et al. 2016) & MIT License & Evaluation \\
          & LKH3(Helsgaun 2017) & Available for academic use & Evaluation \\
          & AM(Kool et al. 2019) & MIT License & Evaluation \\
          & POMO(Kwon et al. 2020) & MIT License & Evaluation \\
          & BOPO(Liao et al. 2025) & MIT License & Evaluation \\
          & POCO(Pan et al. 2025) & MIT License & Evaluation \\
          & MTL-COP(Wang et al. 2025) & MIT License & Evaluation \\
    \bottomrule
    \end{tabular}%
    }
  \label{assets}%
\end{table}%

\end{document}